\documentclass[conference]{IEEEtran}
\IEEEoverridecommandlockouts
\usepackage{cite}
\usepackage{amsmath,amssymb,amsfonts}
\usepackage{algorithmic}
\usepackage{graphicx}
\usepackage{textcomp}
\usepackage{xcolor}
\usepackage{url}

\usepackage{hyperref} 
\usepackage{xurl}

\usepackage[ruled,vlined]{algorithm2e}
\def\BibTeX{{\rm B\kern-.05em{\sc i\kern-.025em b}\kern-.08em
    T\kern-.1667em\lower.7ex\hbox{E}\kern-.125emX}}
\begin{document}

\title{Bayesian Data Augmentation for DNN Retraining with Binomial Outcomes in Vision-Based UAV Landing
}

\author{
    \IEEEauthorblockN{Ashik E Rasul}
    \IEEEauthorblockA{\textit{Department of Mechanical and Nuclear Engineering} \\
    \textit{Tennessee Technological University}\\
    Cookeville, USA \\
    ashik.rasul@outlook.com}
    \and
    \IEEEauthorblockN{Hyung-Jin Yoon}
    \IEEEauthorblockA{\textit{Department of Mechanical and Nuclear Engineering} \\
    \textit{Tennessee Technological University}\\
    Cookeville, USA \\
    hyoon@tntech.edu}
}

\maketitle

\begin{abstract}
In GPS-denied or cluttered urban environments, vision-based landing is essential for reliable UAV missions. Real-world landing sites are often unstructured and highly variable, requiring strong generalization by the perception system. Deep Neural Networks (DNNs) trained with synthetic data augmentation offer a scalable solution for learning landing-site features across diverse vehicle and environmental states. However, computationally expensive DNN retraining, along with challenging performance validation via test flights, limits exhaustive model fine-tuning and necessitates an optimized retraining pipeline. In this work, we deploy a Bayesian data augmentation framework integrated with a photorealistic simulator featuring high-fidelity vehicle dynamics to iteratively retrain the helipad detector DNN, maximizing landing performance as the objective function.

We validate our framework with experiments in a photo-
realistic simulator under different environmental conditions and
vehicle state, demonstrating improved landing performance and tighter confidence intervals on predicted landing outcomes. 

\end{abstract}


\section{Introduction}

\begin{figure*}

    \centering    
    \includegraphics[width=0.95\linewidth]{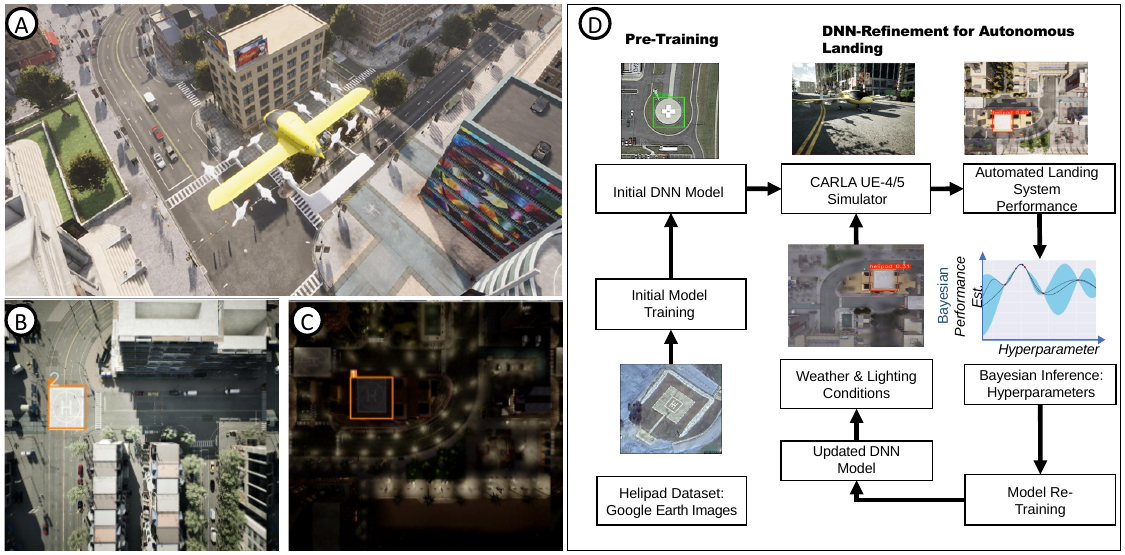}

    \caption{Iterative DNN retraining framework. (A) Photorealistic simulator on CARLA UE4, (B-C) Landing pad detection performance in different lighting conditions, (D) Optimization of landing performance in a Bayesian Optimization loop through Data augmentation and DNN retraining}

    \label{framework}

\end{figure*}

Unmanned Aerial Vehicles (UAVs) have gained significant popularity in recent times for a wide range of military and civilian applications. Traditional navigation methods for UAVs rely on GPS waypoint following, which can be unreliable in cluttered urban environments or provide insufficient accuracy when a high-precision landing is required. Vision-based landing can be a suitable alternative for such scenarios if the landing pad can be reliably detected and tracked throughout the landing maneuver. However, real-world landing sites are often unstructured with variable features, making consistent detection challenging. Deep Neural Networks (DNNs) demonstrate superior performance in learning such complex features compared to traditional computer vision methods, and have been applied in a wide range of perception tasks~\cite{grigorescu2020survey,voulodimos2018deep}. However, training DNNs with random helipad images is not sufficient, as the same helipad may appear in different sizes, shapes, and appearances depending on the vehicle pose, environmental conditions, and camera parameter settings. For real-time object detection and control, general-purpose frameworks such as YOLO (You Only Look Once)~\cite{redmon2016you} are therefore trained with domain-specific datasets~\cite{diwan2023object} with data augmentation to improve the generalization of the model for different sizes, lighting, and aspect ratios of the object. This augmentation process can be validated and refined further based on the vision-based landing performance of the UAV.
However, collecting such landing performance data would involve using expensive aircraft for extended periods to cover varying scenarios, altitude ranges, times of day, and weather conditions.
Photorealistic simulators can substantially reduce this burden by providing controlled, repeatable environments for generating task-specific performance data at scale~\cite{HowTrainDefect2023, dallel2023digital, unity_autonomous_vehicle_training}
Despite these advances, retraining and deployment of DNNs remains a challenging task. 

For example, each iteration of refinement is computationally expensive, requiring substantial resources and time~\cite{thompson2020computational}. This high computational demand often becomes a bottleneck in the optimization process, necessitating improvements in the efficiency of the retraining process itself. The blackbox nature of the DNNs makes the task additionally challenging. Bayesian optimization can be a prominent method for such scenarios, as it systematically searches the decision variable space, balancing exploration and exploitation to identify optimal decision variables without knowing the objective function explicitly~\cite{shahriari2015taking}. By leveraging feedback from simulation outcomes, such as the number of successful landings over a finite number of landing attempts, Bayesian optimization can iteratively refine the model, ensuring effective adaptation to varying vehicle and environmental states.
In this work, we address this challenge by introducing a retraining and deployment framework for perception DNNs, specifically for the autonomous landing operation of UAVs, as illustrated in Figure~\ref{framework}. First, we introduce an iterative DNN retraining process leveraging Bayesian optimization, which optimizes the binomial outcome of the landing attempts (success/failure) by efficiently applying data augmentation to the DNN retraining pipeline.
This process consists of two phases. The first phase evaluates the binomial outcome (landing success/failure) as a performance measure of the vision-guided landing system, given the operation scenarios in a photorealistic simulation. In the second phase, the corresponding performance measure is used in a Bayesian optimization loop that applies the Gaussian Process (GP) prior and Laplace approximation of the posterior to generate the next set of hyperparameters for retraining the model. Finally, the updated model is evaluated in the first phase. These phases close the loop in the DNN refinement process. 

To evaluate our approach, we use a simulation environment~\cite{bansal2025airtaxisim} built in Unreal Engine that models realistic aerial scenarios, including cluttered urban landscapes, varying altitudes, and dynamic lighting. This environment enables thorough testing of the proposed framework, allowing us to study the effect of uncertainty, limited observability, and control feedback in closed-loop settings.

Our key contributions are:

\begin{itemize}    

    \item We present an iterative retraining pipeline for landing pad detection DNN, integrated to a photorealistic simulator, that optimizes the binomial landing outcomes within the simulator as an objective function leveraging Bayesian optimization with GP prior and Laplace approximation of the posterior distribution.

\end{itemize}

\section{Literature Review}
UAVs are being used for a wide range of applications, including terrain mapping~\cite{templetonAutonomousVisionbasedLanding2007}, road following~\cite{frewVisionbasedRoadfollowingUsing2004}, shoreline investigation~\cite{templinApplicationLowCostFixedWing2018}, wildlife monitoring~\cite{olivares2015towards}, and autonomous package delivery~\cite{saundersAutonomousAerialRobotics2024}. Among different UAV configurations, multirotor planes with lift-and-cruise capability have gained popularity in the context of Advanced Air Mobility (AAM)~\cite{johnson2022nasa, silva2018vtol}. These aircraft combine the cruise characteristics of fixed-wing aircraft with the vertical takeoff and landing (VTOL) capability of multirotors. Unlike fixed-wing runway operations, vertical landing requires precise localization of the landing site directly beneath the aircraft. Vision-based perception has emerged as the dominant sensing modality for high-precision autonomous landing in GPS-denied or cluttered environments~\cite{gautam2014survey, xin2022visionlanding}. Despite significant progress, generalized DNN training for diverse landing environments, robustness under lighting and texture variations, and a generalized exploration policy for the landing pad from different UAV poses remain open research directions.

Early vision-based autonomous landing systems relied on cooperative visual markers of specific features, placed on the landing surface. One of the first implementations of such systems demonstrated helicopter landing using image processing-based tracking \cite{saripalli2003vision}. Later, vision-based quadrotor landing was achieved on a moving platform using an onboard camera and nonlinear control \cite{biImplementationAutonomousVisual2013,falanga2017landing}. Vision-based autonomous landing was also demonstrated in GPS-denied environments using customized landing pad design, state estimation, and stabilization~\cite{langeVisionBasedOnboard2009}. These works established the feasibility of marker-based autonomous landing even under motion and disturbances. However, these solutions face challenges in real-world adoption in unknown environments, as the landing markers may not follow a specific feature, shape, or orientation. A Faster Region-based Convolutional Neural Network (Faster R-CNN)-based landing sign identification was proposed to identify the position and direction of the landing sign~\cite{chenIdentificationAutonomousLanding2017}. Later, YOLO~\cite{redmonYOLOv3IncrementalImprovement2018} has been widely integrated for safe landing area detection, pose correction, and navigation of UAVs~\cite{leeLandingSiteInspection2020, symeonidis2021vision}. However, training such perception models for autonomous landing requires large amounts of data collected covering diverse environments and vehicle states. For UAV perception, collecting data at scale can be expensive and unsafe, especially in the early stages of development. Moreover, comprehensive scenarios of varying environmental conditions and vehicle states can be challenging to recreate with real test flights. Photorealistic simulators equipped with high-fidelity vehicle dynamics can substantially reduce this gap by providing task-specific training data~\cite{HowTrainDefect2023, dallel2023digital, unity_autonomous_vehicle_training}. CARLA~\cite{dosovitskiy2017carla} and AirSim \cite{airsim2017} have been widely used for such synthetic data generation and control validation. However, the simulation-to-real gap still remains a major challenge for UAV perception model deployment, which can be further reduced by synthetically augmenting the training data itself through careful tuning of training parameters such as scale and brightness. Though such fine-tuning enables the DNNs to generalize well across varying altitude and lighting conditions, each iteration of retraining and validation is computationally expensive, which motivates the use of Bayesian optimization to efficiently search the hyperparameter space for optimal performance~\cite{shahriari2015taking}. Bayesian optimization based on stochastic binary outcomes is successfully demonstrated for different robotics applications~\cite{tesch2013expensive, rofer2024bayesian}. However, to the best of our knowledge, none of the prior works optimized perception DNNs for the autonomous landing task with a binomial simulation outcome.
In this work, we place a Gaussian prior over the true unknown function and adopt a binomial observation model. We deploy the Laplace approximation method to approximate the non-Gaussian posterior. Finally, we leverage upper confidence bound (UCB) as an acquisition function to explore the hyperparameter space~\cite{nickisch2008approximations}.

\section{Problem Formulation}

The Bayesian optimization in this work aims to maximize the landing success rate within a defined altitude range in varying lighting conditions by iteratively retraining the object detection DNN (YOLO). We consider scale augmentation factor (S) and brightness augmentation (B) as relevant hyperparameters and define them as decision variables: $\textbf{x}=[S,B]^\top$. The objective is to identify the configuration that maximizes the probability of a successful landing, $p(\textbf{x})$: 

\begin{equation}
    \textbf{x}^*=\arg\max _{\textbf{x}\in \mathcal{X}} p(\textbf{x})
\end{equation}
Here, $\mathcal{X}$ denotes the bounded hyperparameter space. 
The true landing performance, $f(\textbf{x}):\mathcal{X}\to \mathbb{R}$, is a latent function that cannot be observed directly. Instead, for each configuration of $\textbf{x}$, we perform $m_t$ independent landing trials and observe $y_i \in \{0, 1, \dots, m_t\}$ successful landings. The successful landing probability is related to the latent function through the probit link function:
\begin{equation}
    p(\textbf{x})=\Phi(f(\textbf{x}))
\end{equation}
Here, $\Phi(\cdot)=\int_{-\infty}^t \mathcal{N}(\tau|0,1)d\tau$ is the cumulative distribution function of the standard normal distribution that works as a link function. The link function maps the latent landing performance $f(\textbf{x})\in \mathbb{R}$ to the probability space $p(\textbf{x})\in[0,1]$. Given the latent value $f_i = f({\textbf{x}}_i)$, the 
$y_i$ successes out of a fixed number of trials ($m_t$) follow a Binomial distribution:
\begin{equation}
    P(y_i \mid f_i)
    = \binom{m_t}{y_i}
      \Phi(f_i)^{y_i}\,
      \bigl(1 - \Phi(f_i)\bigr)^{m_t - y_i}
\end{equation}
The dataset after $n$ queries is therefore
$\mathcal{D}_n = \{\textbf{x}_i,\, y_i\}_{i=1}^{n}$,
where each entry records the queried configuration and the observed
success count.

We model the unknown latent function $f(\mathbf{x})$ as a Gaussian Process (GP):
\begin{equation}
    f(\mathbf{x}) \sim \mathcal{GP}\bigl(m(\mathbf{x}),\,
    k(\mathbf{x}, \mathbf{x}')\bigr)
\end{equation}
The mean function is set to $m(\mathbf{x}) = 0$ as no prior information about the landing performance function is available. For the covariance function, we adopt the Radial Basis Function (RBF) kernel, also known as the squared exponential kernel:
\begin{equation}
    k(\mathbf{x}, \mathbf{x}')
    = \sigma_f^2
      \exp\!\left(
        -\frac{\|\mathbf{x} - \mathbf{x}'\|^2}{2\ell^2}
      \right)
\end{equation}
where $\sigma_f^2$ is the signal variance and $\ell$ is the length-scale. The RBF kernel assumes that nearby configurations in scale and brightness augmentation in $[0,1]^2$ produce similar detector performance with no abrupt transitions. As both the scale and brightness have the same operating range for data augmentation, we assume a single isotropic length-scale $\ell$. Given observations $\mathcal{D}_n$, the posterior over the latent function values $\mathbf{f} = [f_1, \dots, f_n]^\top$ is:
\begin{equation}
    p(\mathbf{f} \mid \mathcal{D}_n)
    \propto
    p(\mathbf{y} \mid \mathbf{f})\,
    p(\mathbf{f} \mid \mathbf{X})
\end{equation}
The Binomial-Probit likelihood, $p(\mathbf{y} \mid \mathbf{f})$, makes this posterior non-Gaussian and analytically intractable. We adopt the Laplace approximation, which fits a Gaussian at the mode of the log-posterior:
\begin{equation}
    p(\mathbf{f} \mid \mathcal{D}_n)
    \approx \mathcal{N}(\hat{\mathbf{f}},\, \boldsymbol{\Sigma})
\end{equation}
where $\hat{\mathbf{f}}$ is the MAP estimate found by Newton
iterations on the log-posterior, and the posterior covariance is:
\begin{equation}
    \boldsymbol{\Sigma} = \left(K^{-1} + W\right)^{-1}
\end{equation}
Here $K$ is the $n \times n$ kernel matrix with entries $K_{ij} = k(\mathbf{x}_i, \mathbf{x}_j)$, and $W$ is a diagonal matrix of negative second derivatives of the log-likelihood evaluated at $\hat{\mathbf{f}}$~\cite{nickisch2008approximations}. With the number of Binomial trials $m_t$ fixed, the diagonal entries are:
\begin{equation}
\begin{split}
    W_{ii}
    &=
    \phi^2(\hat{f}_i)
    \left[
      \frac{y_i}{\Phi(\hat{f}_i)^2}
      +
      \frac{m_t - y_i}{\bigl(1 - \Phi(\hat{f}_i)\bigr)^2}
    \right]\\
    +
    & \quad \hat{f}_i\,\phi(\hat{f}_i)
    \left[
      \frac{y_i}{\Phi(\hat{f}_i)}
      -
      \frac{m_t - y_i}{1 - \Phi(\hat{f}_i)}
    \right]
    \end{split}
\end{equation}
where $\phi(\cdot)$ is the standard normal PDF. 
%
%
Now, for a new candidate configuration $\mathbf{x}_*$, the predictive distribution over the latent function is approximately Gaussian:
\begin{equation}
    p(f_* \mid \mathbf{x}_*,\, \mathcal{D}_n)
    \approx
    \mathcal{N}\!\left(\mu(\mathbf{x}_*),\,
    \sigma^2(\mathbf{x}_*)\right)
\end{equation}
with posterior predictive mean and variance:
\begin{equation}
\begin{split}
    \mu(\mathbf{x}_*)
    =& \mathbf{k}_*^\top K^{-1} \hat{\mathbf{f}}, \\
    \qquad
    \sigma^2(\mathbf{x}_*)
    =& k(\mathbf{x}_*, \mathbf{x}_*)
      - \mathbf{k}_*^\top
        \!\left(K^{-1} - K^{-1}\boldsymbol{\Sigma} K^{-1}\right)
        \mathbf{k}_*
\end{split}
\end{equation}
where $\mathbf{k}_* = [k(\mathbf{x}_*, \mathbf{x}_1), \dots, k(\mathbf{x}_*, \mathbf{x}_n)]^\top$. The estimated landing success probability at $\mathbf{x}_*$ is obtained by marginalizing the probit link over the predictive latent distribution, yielding the moment-matched approximation:
\begin{equation}
    \bar{p}(\mathbf{x}_*)
    = \mathbb{E}\left[\Phi(f_*)\right]
    \approx
    \Phi\!\left(
      \frac{\mu(\mathbf{x}_*)}{\sqrt{1 + \sigma^2(\mathbf{x}_*)}}
    \right)
\end{equation}

\begin{algorithm}
\caption{Bayesian Optimization for Scale and Brightness}
\label{alg:bayesian_optimization}
\KwIn{Parameter space $\mathcal{X} = \{(S, B) : S, B \in [0,1]\}$,
    GP prior, UCB,
    initial points $n_0$, budget $T$, tolerance $\epsilon$}
\KwOut{Optimal parameters $(S^*, B^*)$}

\tcp{Initialization}
$\mathcal{X}_0 \leftarrow \{\mathbf{x}_i\}_{i=1}^{n_0}$ uniformly spaced over 
$\mathcal{X}$\;
\For{$\mathbf{x}_i \in \mathcal{X}_0$}{
    $M_i \leftarrow \mathrm{TrainModel}(\mathbf{x}_i)$\;
    $y_i \leftarrow \mathrm{EvaluateModel}(M_i,\, m_t)$\;
}
$\mathcal{D}_0 \leftarrow \{(\mathbf{x}_i,\, y_i)\}_{i=1}^{n_0}$\;

\BlankLine
\tcp{Bayesian optimization loop}
\For{$n = 0$ \KwTo $T$}{
    Update GP posterior using $\mathcal{D}_n$ (Laplace approximation)\;
    \BlankLine
    $(S_\mathrm{next}, B_\mathrm{next}) \leftarrow 
    \arg\max_{(S,B) \in \mathcal{X}}\; \mathrm{UCB}(S, B \mid \mathcal{D}_n)$\;
    \BlankLine
    $M \leftarrow \mathrm{TrainModel}(S_\mathrm{next},\, B_\mathrm{next})$\;
    $y_\mathrm{next} \leftarrow \mathrm{EvaluateModel}(M,\, m_t)$ 
    \BlankLine
    $\mathcal{D}_{n+1} \leftarrow \mathcal{D}_n \cup 
    \{(S_\mathrm{next},\, B_\mathrm{next},\, y_\mathrm{next})\}$\;
    \BlankLine
    \If{$\max_{\mathbf{x} \in \mathcal{X}}\; 
        \mathrm{UCB}(\mathbf{x};\, \mathcal{D}_n) < \epsilon$}{
        \textbf{break} 
    }
}

\BlankLine
\Return $(S^*,\, B^*) = \arg\max_{(S,B) \in \mathcal{X}_{0:n}}\; \bar{p}(S,\, B)$\;
\end{algorithm}

This quantity is the GP's current best estimate of the landing success rate at any untried configuration. To select the next configuration to query, we use the UCB acquisition function defined on the latent GP:
\begin{equation}
    \mathrm{UCB}(\mathbf{x})
    = \mu(\mathbf{x}) + \kappa\, \sigma(\mathbf{x})
\end{equation}
\begin{equation}
    \mathbf{x}_{n+1}
    = \arg\max_{\mathbf{x} \in \mathcal{X}}\
    \mathrm{UCB}(\mathbf{x})
\end{equation}
Here, $\kappa$ is the tunable parameter that trades off between exploration and exploitation. 
After observing $y_{n+1}$ successes at $\mathbf{x}_{n+1}$, the
dataset is updated as:
\begin{equation}
\mathcal{D}_{n+1} = \mathcal{D}_n \cup
  \left\{(\mathbf{x}_{n+1},\, y_{n+1})\right\}
\end{equation}
Since the GP is a non-parametric model, updating the dataset corresponds to updating the GP itself. The process is repeated until the evaluation budget is exhausted or a stopping condition is met as:
\begin{equation*}
    \text{Stop if: } \max_{\mathbf{x} \in \mathcal{X}} \mathrm{UCB}(\mathbf{x};\, \mathcal{D}_n) < \epsilon,
\end{equation*}
where $\epsilon$ is a predefined threshold~\cite{snoek2012practical}. The iterative model refinement process is summarized in Algorithm~\ref{alg:bayesian_optimization}.

\section{Experiments and Results}


\subsection{Photorealistic Simulator and Vehicle Dynamics}
The photorealistic simulator~\cite{bansal2025airtaxisim} used in this work is developed using the existing autonomous car simulation application, CARLA~\cite{dosovitskiy2017carla}. The simulator integrates vehicle dynamics, controller, planning, and perception.


\

The simulator also allows us to define many environmental conditions, such as different landing landscapes, weather, and lighting, enabling us to generate realistic, challenging scenarios, as shown in Figure~\ref{framework}. This flexibility enables a safe, repeatable testing environment without the risks associated with real-world UAV operations, especially in the early stages of development. 
The simulator leverages Generic Urban Air Mobility (GUAM)~\cite{GUAM2024}, developed by NASA, to simulate vehicle dynamics.
GUAM can simulate common rigid-body 6-DOF (Degree of Freedom) frames of reference, which include a wide range of aerospace signals and quantities of interest. The simulator's modular architecture readily supports the integration and replacement of aircraft models, sensors, actuator models, and control algorithms. GUAM was later migrated from MATLAB Simulink to a Python environment using Google’s JAX framework~\cite{jaxguam2023}. JAX leverages GPU parallel computing and provides automatic differentiation, similar to PyTorch~\cite{NEURIPS2019_9015}, enabling faster simulation speeds of up to $200\,\text{Hz}$. To integrate CARLA with JAX-GUAM and the perception module, ROS~\cite{ros} is used as a middleware. 



\subsection{Landing Controller}
\begin{figure}
        \centering       
        \includegraphics[width=0.35\textwidth]{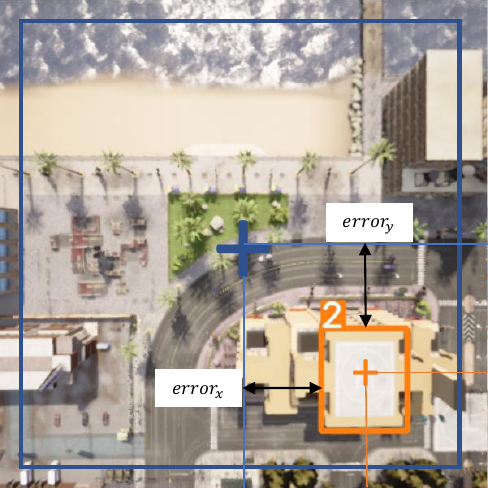}
        \caption{Error signal for the controller, the observed bounding box (orange) is compared with the target bounding box(blue) within the FOV.}
        \label{fig:error}
\end{figure}

We assume that if the landing pad is positioned directly under the UAV, the detected bounding box will be placed at the center of a target area within the FOV at the world frame. The detected bounding box is marked as orange, and the target area is marked as blue, as shown in Figure~\ref{fig:error}. These two areas generate an error vector that can be used to drive the UAV to the target location. 
To handle object detection of a moving target, we implement the SORT (Simple Online and Real-time Tracking) method~\cite{bewley2016sort}. It leverages a Kalman filter for consistent tracking of the correct helipad bounding box in dynamic environments, even when the target is partially occluded or temporarily lost. With accurate object detection and SORT, the UAV can continuously produce the necessary error signal for the control algorithm.

We define the landing control error as follows: ${e}={b}_c - {b}_o$,
where ${b}_c$ is the target area and ${b}_o$ is the object area. Both the ${x}$ and ${y}$ components of the error can be found from the center deviation of the areas in the respective axes. For the ${z}$ component of the error, we take their size difference. This error is high when the UAV is at a high altitude and is low closer to the landing pad altitude. We assume that the object area is less than the target area throughout the descent, and they overlap at the ideal landing position. 
Without loss of generality and for simplicity, we consider a PID controller~\cite{aastrom2021feedback} to compute the corrective velocity vector from the error signals, ${v}_{\text{corr}}=[v_x, v_y, v_z]^\top$:
\begin{equation}
   v_i = K_p e_i + K_i \int e_i \, dt + K_d \frac{de_i}{dt}, \quad i \in \{x, y, z\}, 
\end{equation}
where \(v_i\) is the corrective velocity along axis \(i\), \(e_i\) is the error along axis \(i\), and \(K_p\), \(K_i\), \(K_d\) are the proportional, integral, and derivative gains for each respective axis~\cite{csumnu2017simulation}.
The calculated velocity commands are constrained to prevent unmanageable overshoot and then fed into the vehicle dynamics module, which updates the vehicle's state, guiding it closer to the ideal landing position based on the corrective velocity vector $v_i$.

\subsection{Dataset Processing and Initial Model Training}

The first step of our experiment focuses on data preprocessing and augmentation to train reproducible image detection DNNs, specifically YOLOv8~\cite{terven2023yolo}. As a dataset, we use Google Earth images of helipads~\cite{bitoun2020helipadcat}. 

By default, YOLO models employ random data augmentations during training to enhance model robustness and generalization. These augmentations include techniques like random resizing, flipping, cropping, color, and brightness adjustments. The randomness inherent in these augmentations leads to non-reproducible results across different training runs~\cite{ultralytics2024}. To achieve reproducibility, we turn off the random augmentation of YOLO training, which gives us control over the image augmentation during the training cycle. 
This also allows us to isolate the effects of specific hyperparameters and observe the model's behavior on unseen data. However, this method of hyperparameter tuning has the inherent risk of overfitting the model that performs well only on that specific scenario, reducing its ability to generalize to unseen data.
We reduce the risk of overfitting by validating model performance while randomly varying the UAV offset from the landing pad within a wide range of altitudes. 



\begin{figure}
    \centering
        \includegraphics[width=0.35\textwidth]{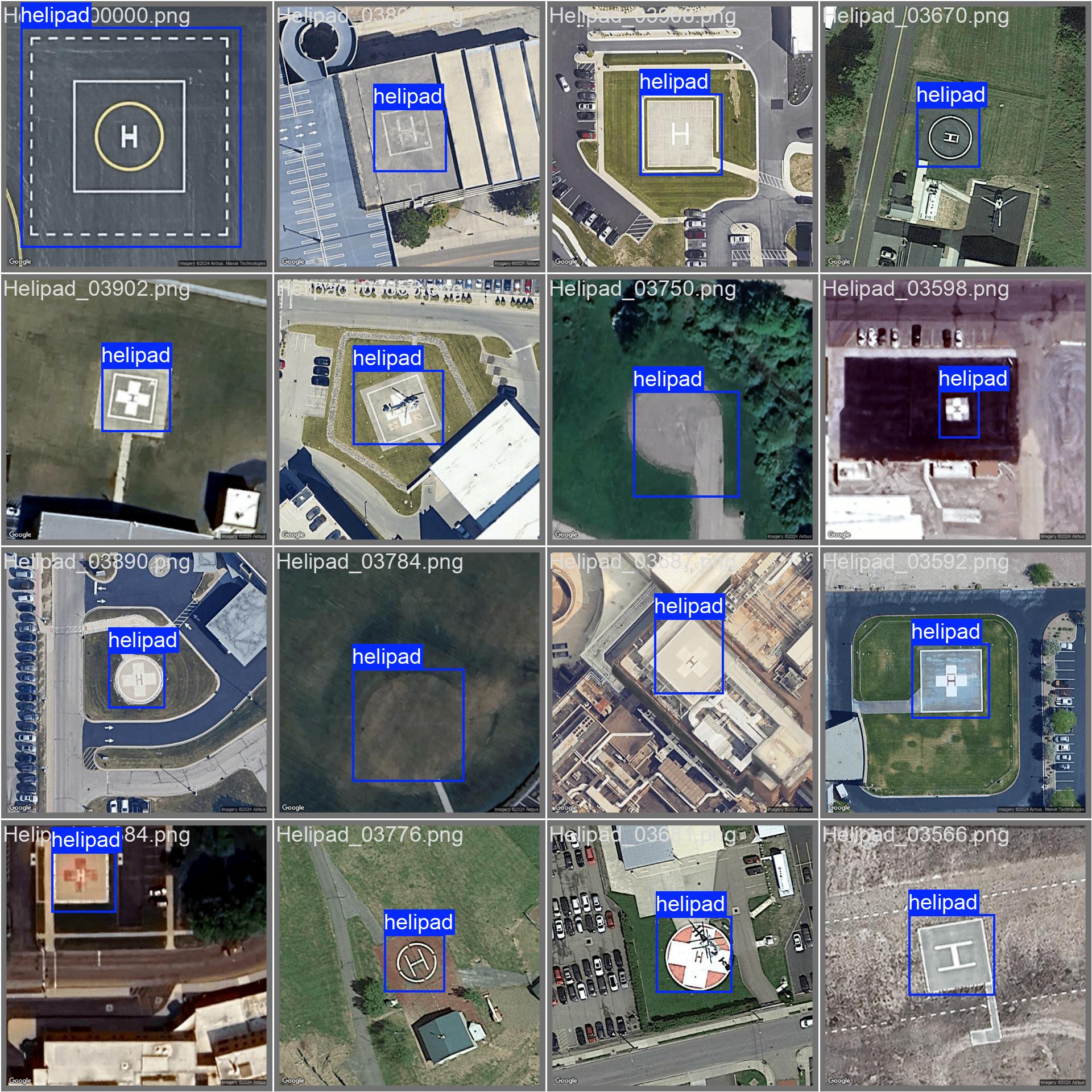}
        \caption{Helipad Dataset~\cite{bitoun2020helipadcat} showing Google Earth images of different helipads with their respective bounding box annotations.}
        \label{fig:helipad}
\end{figure}

\begin{figure}
    \centering
    \includegraphics[width=.95\linewidth]{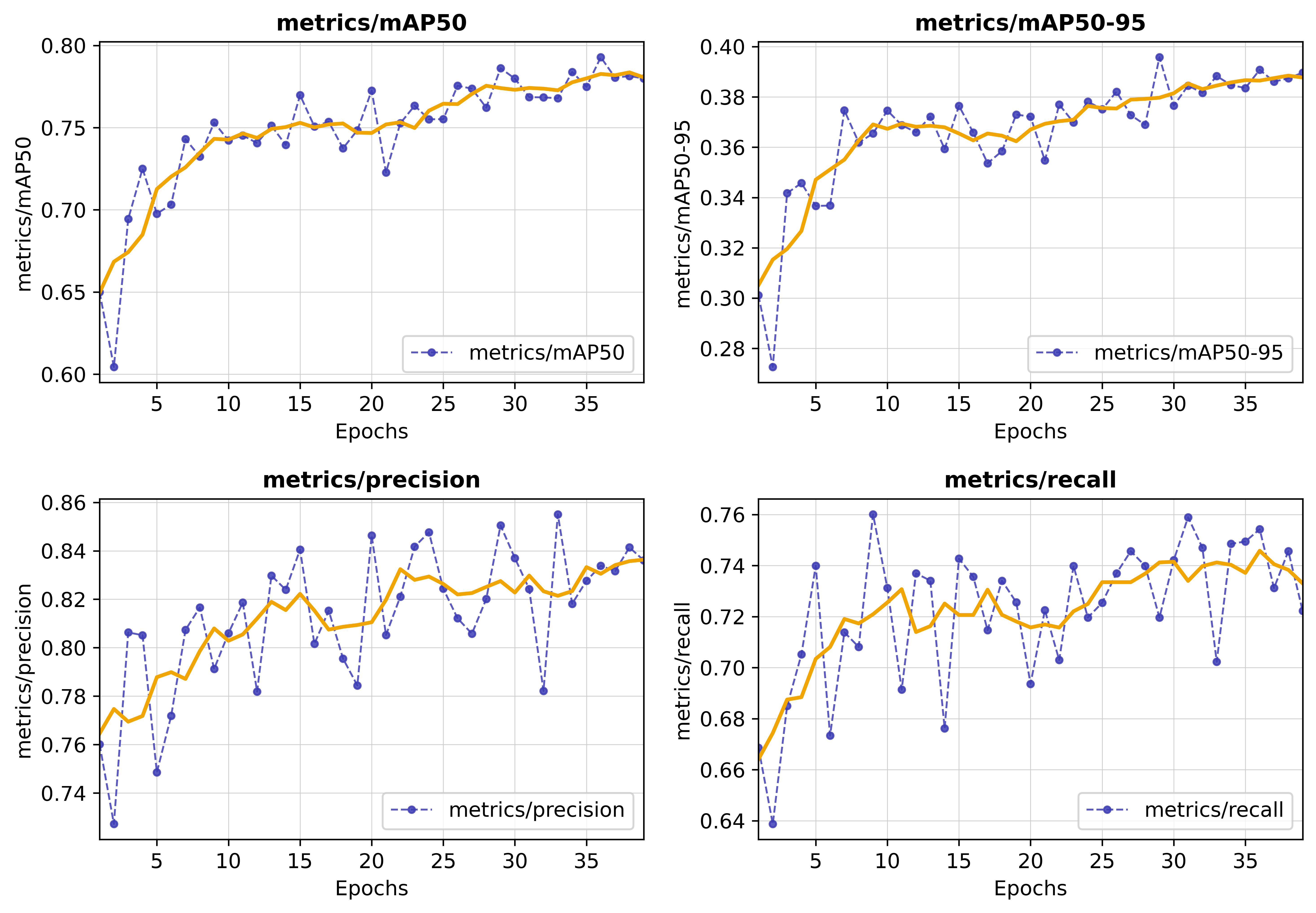}
    \caption{Representative single DNN training result} 
    \label{fig:result}
\end{figure}

\begin{figure*}
    \centering
    \includegraphics[width=0.99\linewidth]{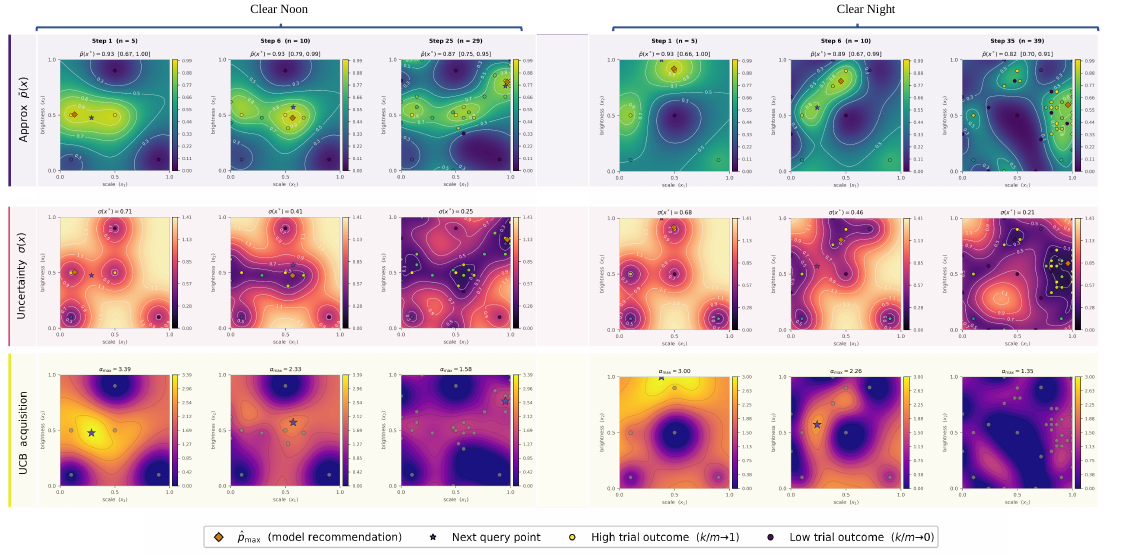}
    \caption{GP posterior and acquisition function representation at $3$ different iteration steps (\texttt{CLEAR-NOON} and \texttt{CLEAR-NIGHT} weather in CARLA). The top row shows the landing success rate probability $\bar{p}(x)$ with $95\%$ CI, the middle row shows the uncertainty-$\sigma(x)$ for the latent function, and the bottom row shows the respective UCB acquisition function values.}
    \label{fig:bo_result}
\end{figure*}

From the dataset, 80\% of the images are randomly selected for training, while the remaining 20\% are reserved for validation. Figure~\ref{fig:helipad} shows sample training images annotated with the helipad class bounding boxes, demonstrating the diversity and distribution of the dataset. The YOLO model outputs bounding box coordinates with confidence scores, indicating the likelihood of correct helipad detection. The model’s performance is evaluated using standard object detection metrics, including mean Average Precision (mAP) at IoU (Intersection over Union) thresholds of 50\% (mAP50) and a range from 50\% to 95\% (mAP50-95), as shown in Figure~\ref{fig:result}. These metrics provide a comprehensive assessment of the model's detection accuracy. IoU measures the overlap between the predicted bounding box and the ground truth bounding box, serving as a criterion for correct detection~\cite{lin2014microsoft}. Precision at a single IoU threshold (50\%) is measured by mAP50, indicating how well the model detects objects at a basic level of overlap, and mAP50-95 averages precision over multiple IoU thresholds~\cite{lin2014microsoft}. This metric mAP50-95 provides a more rigorous evaluation of the model's performance, especially for scenarios requiring fine-grained object detection~\cite{lin2014microsoft}. The curves in the top two charts illustrate that as training progresses over epochs, both mAP50 and mAP50-95 steadily increase, indicating improvements in the model's ability to detect helipads accurately. The precision and recall metrics, shown in the bottom two charts, provide further insight into the model’s performance. Precision evaluates the proportion of correctly detected helipads out of all detections made, reflecting the model’s ability to avoid false positives~\cite{lin2014microsoft}. Recall measures the proportion of correctly detected helipads out of all actual helipads, indicating how well the model avoids false negatives~\cite{lin2014microsoft}. These trends highlight the model's increasing generalization capability and robustness when applied to the validation set. 

\subsection{Landing Performance Evaluation}

We train five initial DNN models using uniformly spaced augmentation values (scale and brightness). The models are then integrated into the simulation environment to assess landing success rates across 10 randomly generated initial positions within a $120\,\text{m}$ altitude range, divided into 2 altitude levels. The initial positions are also generated with a random offset in the horizontal plane within a range of $\pm100\,\text{m}$, depending on the altitude. A landing is considered successful if the vehicle reaches within $4\,\text{m}$ of the ideal center of the landing pad along the horizontal plane and within $\pm3\,\text{m}$ of the ideal center of the landing pad in altitude. We select a \texttt{CLEAR-NOON} weather condition in CARLA as our initial model refinement scenario. The UAV camera resolution is set at $640\times 640$ pixels along with other default parameters. The initial models achieve varying outcomes ranging from $0$ to $10$ landing successes. These evaluation data points, along with their respective scale and brightness augmentation values, are fed into the Bayesian optimization algorithm that suggests the next iteration hyperparameter.



\subsection{Bayesian Optimization for Iterative Retraining}

Each landing trial produces a binary outcome: the UAV either lands within the success threshold or it does not. As we are considering scale and brightness augmentation for this work, the search space becomes a two-dimensional hyperparameter domain, with axes corresponding to the data augmentation parameters: image scale factor and HSV brightness (\texttt{hsv\_v}). The RBF kernel is initialized with length-scale $\ell = 0.20$ and output variance $\sigma_f^2 = 2.0$. Because the Binomial-Probit likelihood is non-Gaussian, the GP posterior is analytically intractable. We use the Laplace approximation to obtain a Gaussian surrogate at each iteration.
For the UCB acquisition function, we set the exploration weight $\kappa = 1.5$ and a total exploration budget of 30, including 5 initial space-filling points.

Early in the search, although we receive some high simulation outcomes, the GP posterior recommendation carries substantial uncertainty. At the first active step, we get $\bar{p}_{\max} = 0.93$ with a wide 95\% credible interval (CI) of $[0.67, 1.00]$. As we conduct more active queries with the acquisition function, the posterior uncertainty contracts considerably. After 25 active iterations, the model's recommendation converges to a configuration near $(x_1, x_2) \approx (0.97, 0.80)$, with $\bar{p}_{\max} = 0.87$ and a 95\% CI of $[0.75, 0.95]$. This narrower CI indicates that the surrogate model has accumulated sufficient evidence to characterize the landing success probability at the recommended configuration with considerably greater confidence.

We repeat the DNN refinement process for the \texttt{CLEAR-NIGHT} condition in CARLA, considering the scale and brightness as augmentation parameters. We also follow the same initialization for the RBF kernel and set $\kappa=1.5$. At the first active step, we get $\bar{p}_{\max} = 0.93$ with a wide 95\% credible interval (CI) of $[0.66, 1.00]$. We observe a slower convergence of the maximum value of the UCB acquisition function governed by sparse landing successes under night conditions. Therefore, we increase the active query budget, and after 35 active iterations, the model's recommendation converges to a configuration near $(x_1, x_2) \approx (0.95, 0.60)$, with $\bar{p}_{\max} = 0.82$ and a 95\% CI of $[0.70, 0.91]$. 

\begin{figure}
    \centering
    \includegraphics[width=0.75\linewidth]{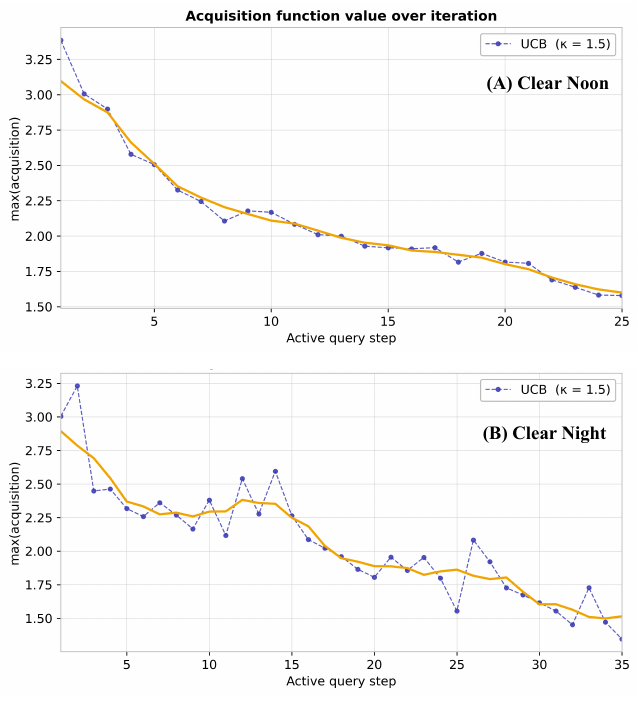}
    \caption{Evolution of the UCB acquisition function values over the active queries. Results are for \texttt{CLEAR-NOON} (Top) and \texttt{CLEAR-NIGHT} (Bottom) weather in CARLA. }
    \label{fig:acq_ucb}
\end{figure}

The evolution of the GP posterior ($\bar{p}(\textbf{x})$), uncertainty ($\sigma(\textbf{x})$), and UCB acquisition surface at three representative iterations is shown in Figure~\ref{fig:bo_result}, illustrating how the surrogate progressively concentrates probability mass in the high-performance region of the augmentation space. The convergence of the acquisition function value is shown in Figure~\ref{fig:acq_ucb}.

\section{Conclusion}

In this work, we developed a framework for iteratively retraining a perception DNN by augmenting the training dataset using landing performance feedback from a photorealistic simulator. Though we considered scale and brightness augmentation for optimizing the DNN model, the Bayesian optimization can be further extended to include any other data-augmentation parameters, depending on the deployment environment. For handling binomial outcomes of the landing performance, we considered Laplace approximation methods. This can be further extended to other approximation methods such as Expectation Propagation~\cite{minka2013expectation} and Markov Chain Monte Carlo~\cite{gilks1995markov}. As the complexity of the deployment environment and the data-augmentation pipeline increases, the model can be further tuned with different choices of kernel functions and parameters.

\bibliographystyle{ieeetr}
\bibliography{bibliography}

\end{document}